\documentclass[a4paper]{llncs}
\usepackage{booktabs}  
\usepackage[pagebackref=false,breaklinks=true,letterpaper=true,colorlinks,bookmarks=false]{hyperref}
\usepackage{amssymb}
\usepackage{graphicx}
\usepackage{array}
\usepackage{url}

\begin{document}

\title{A Pulmonary Nodule Detection Model Based on Progressive Resolution and Hierarchical Saliency}

\author{Junjie Zhang$^{1}$,
    Yong Xia$^{1,2*}$,
    and Yanning Zhang$^{1}$}

\institute{$^1$ Shaanxi Key Lab of Speech \& Image Information Processing (SAIIP), School of Computer Science, Northwestern Polytechnical University, Xi'an 710072, China  \\
$^2$Centre for Multidisciplinary Convergence Computing (CMCC), School of Computer Science and Technology, Northwestern Polytechnical University, Xi'an 710072, China \\
Corresponding Author's Email: yxia@nwpu.edu.cn
}


\maketitle

\begin{abstract}
Detection of pulmonary nodules on chest CT is an essential step in the early diagnosis of lung cancer, which is critical for best patient care. Although a number of computer-aided nodule detection methods have been published in the literature, these methods still have two major drawbacks: missing out true nodules during the detection of nodule candidates and less-accurate identification of nodules from non-nodule. In this paper, we propose an automated pulmonary nodule detection algorithm that jointly combines progressive resolution and hierarchical saliency. Specifically, we design a 3D progressive resolution-based densely dilated FCN, namely the progressive resolution network (PRN), to detect nodule candidates inside the lung, and construct a densely dilated 3D CNN with hierarchical saliency, namely the hierarchical saliency network (HSN), to simultaneously identify genuine nodules from those candidates and estimate the diameters of nodules. We evaluated our algorithm on the benchmark LUng Nodule Analysis 2016 (LUNA16) dataset and achieved a state-of-the-art detection score. Our results suggest that the proposed algorithm can effectively detect pulmonary nodules on chest CT and accurately estimate their diameters.
\end{abstract}

\section{Introduction}
Lung cancer is the leading cause of all cancer-related deaths for both men and women \cite{Ferlay2015Cancer}. The 5-year survival rate of lung cancer patients is only about 16\% on average; this number, however, can reach approximately 54\% if the diagnosis is made at an early stage of the disease \cite{Baldwin2015Prediction}. Since malignant pulmonary nodules may be primary lung tumors or metastases, early detection of pulmonary nodules is critical for best patient care \cite{Kohan2013N}.
On chest CT scans, a pulmonary nodule usually refers to a “spot” of less than 3 cm in diameter on the lung \cite{Callister2016How}. Many automated pulmonary nodule detection methods have been proposed in the literature.  Most of them extract hand-crafted features in each suspicious lesion and train a classifier to determine if the lesion is a pulmonary nodule or not \cite{Valente2016Automatic}. Nithila et al. \cite{Nithila2016Automatic} adopted the intensity cluster, rolling ball and active contour model (ACM) based algorithms to detect solidary, juxta-pleural and juxta-vascular nodules, respectively, then calculated statistical and texture features of those nodules, and used a back propagation neural network (BPNN) optimized by the particle swarm optimization (PSO) technique to improve the detection. Wu et al. \cite{Wu2016Correlation} first applied the thresholding, region growing and morphology operations to segment pulmonary nodules, then extracted 34 visual features on each nodule, and employed a support vector machine (SVM) to eliminate false positive ones. Despite their prevalence, these methods may suffer from limited accuracy, due to the intractable optimization of hand-crafted feature extraction and classification.
 Recently, deep learning techniques have been widely applied to many medical image analysis tasks, including pulmonary nodule detection. They have distinct advantages over traditional methods in the joint representation learning and pattern classification in a unified network, and hence are able to largely address the drawbacks of traditional methods. Hamidian et al. \cite{Hamidian20173D} used a 3D fully convolutional network (FCN) to generate a score map for the identification of nodule candidates and employed another 3D deep convolutional neural network (DCNN) for nodule and non-nodule classification. Dou et al. \cite{Dou2017Automated} also implemented a FCN to produce nodule candidates, and incorporated two residual blocks into a 3D DCNN for nodule detection. Although such computeraided detection (CADe) systems have yielded promising results, they still have two major drawbacks: (a) previous FCN-like deep neural networks usually detect fewer nodule candidates, and this reduces the complexity of the subsequent nodule and non-nodule classification but may lead to missing out true nodules; and (b) using traditional 3D DCNN to differentiate nodules from non-nodule tissues is still less-accurate, since the visual difference between them is extremely subtle.
To address both drawbacks, we propose a novel algorithm to detect pulmonary nodules on chest CT scans. The main contributions of this work include: (a) designing a 3D progressive resolution-based densely dilated FCN called the progressive resolution network (PRN) to detect nodule candidates inside the lung, and (b) designing a densely dilated 3D CNN with hierarchical saliency, namely the hierarchical saliency network (HSN), to simultaneously identify genuine nodules from those candidates and estimate the diameters of nodules. We have evaluated the proposed method on the benchmark LUng Nodule Analysis 2016 (LUNA16) dataset \cite{Aaa2016Validation} and achieved higher performance than the top five records on the LUNA16 Challenge leaderboard.

\section{Dataset}
The LUNA16 dataset \cite{Aaa2016Validation} used for this study contains 888 chest CT scans and 1186 pulmonary nodules. Each scan, with the slice thickness less than 2.5 mm and slice size of 512$\times$512 voxels, and was annotated during a two-phase procedure by four experienced radiologists. Each radiologist marked lesions they identified as non-nodule, nodule $<$ 3 mm, and nodules $>=$ 3 mm. The reference standard of the LUNA16 challenge consists of all nodules $>= 3$ mm accepted by at least 3 out of 4 radiologists.

\section{Method}
\begin{figure}
\setlength{\intextsep}{0pt plus 0pt minus 1pt}
\setlength{\belowdisplayskip}{-1pt}
\setlength{\abovedisplayskip}{-1pt}
\setlength{\abovecaptionskip}{-1pt}
\setlength{\belowcaptionskip}{-1pt}
\centering
\includegraphics[width=0.90\textwidth]{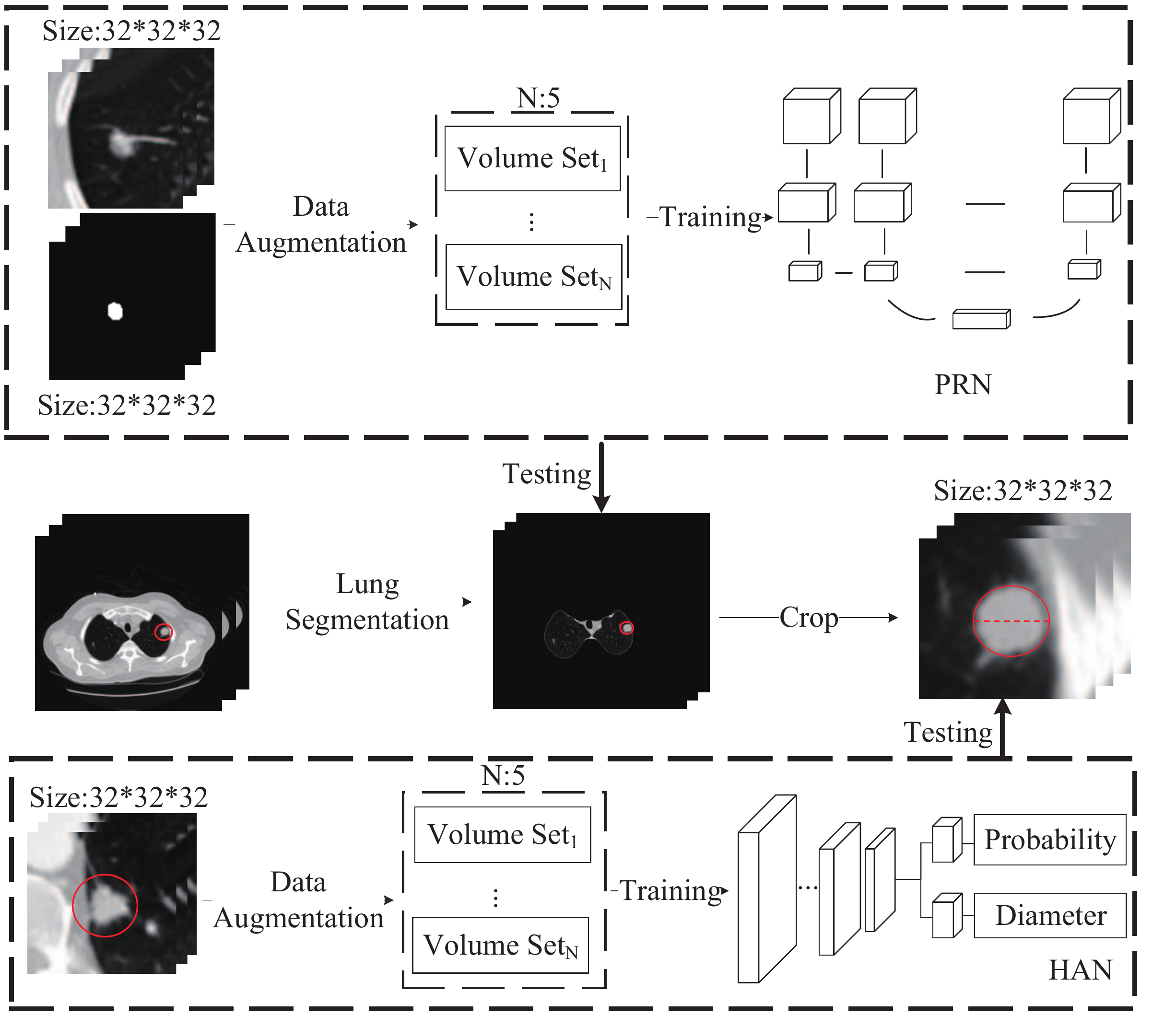}
\caption{Diagram of the proposed pulmonary nodule detection method}
\label{fig1}
\end{figure}

The proposed pulmonary nodule detection algorithm (see Fig. \ref{fig1}) consists of three steps: lung segmentation, PRN-based detection of nodule candidates and HSN-based identification of genuine nodules from those candidates. We now delve into each of the three steps.

\subsection{Data Resampling and Lung segmentation}

To normalize the variable spatial resolution of chest CT scans, we resampled each scan to the uniform voxel size of 1.0$\times$1.0$\times$1.0mm${^3}$. Before detecting nodule candidates, it is also necessary to segment the lung, since outside-lung organs and tissues such as sternums may cause an extremely adverse effect on the detection. The segmentation process include: (a) binarizing each resampled CT scan using the OTSU algorithm \cite{Ohtsu1979A} on a slice-by-slice basis; (b) using the morphological closing to fill holes and the morphological dilation with a disk structure element of size 5 to produce a lung mask that covers as much lung tissues as possible; and (c) applying this mask to the resampled CT scan to get the volume of lung.

\subsection{PRN-based Detection of Nodule Candidates}
\begin{figure}
\setlength{\intextsep}{0pt plus 0pt minus 1pt}
\setlength{\belowdisplayskip}{-1pt}
\setlength{\abovedisplayskip}{-1pt}
\setlength{\abovecaptionskip}{-1pt}
\setlength{\belowcaptionskip}{-1pt}
\centering
\includegraphics[width=0.90\textwidth]{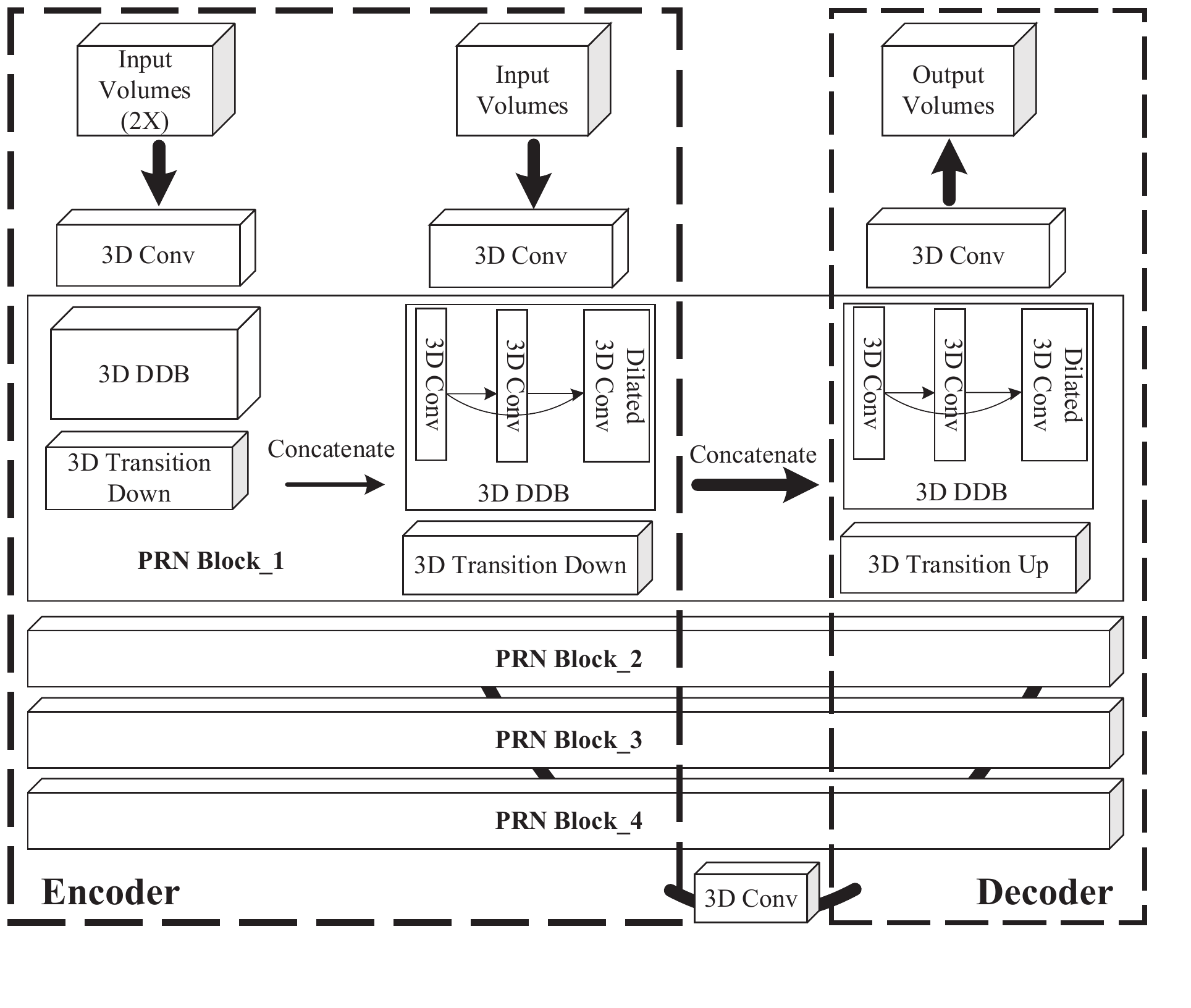}
\caption{Diagram of the proposed pulmonary nodule detection method}
\label{fig2}
\end{figure}
We slid a 32$\times$32$\times$32 window over the obtained lung volume and extracted the volumetric patch inside the window at each location as the input of PRN for the detection of nodule candidates.
The proposed PRN model (see Fig. 3) consists of a dual-path encoder and a decoder. The encoder extracts and transmits features by aggressive resolution, and the decoder produces nodule candidates. We first enlarge each input volumetric patch to 64$\times$64$\times$64, and then feed the patch and its enlarged copy to two encoder paths, respectively. Each encoder path starts with a 3$\times$3$\times$3 convolution layer with the rectified linear unit (ReLU) activation function, followed by four PRN blocks. Each PRN block is composed of a 3D densely dilate block (DDB) that is similar to the dense block \cite{Huang2017Densely} and a transition down unit. The DDB contains three convolution layers with the kernel size of 3$\times$3$\times$3, 5$\times$5$\times$5 and 3$\times$3$\times$3, respectively, and a dilation rate of 2, which can reduce the complexity and aggregate multiscale contextual information without losing resolution \cite{Yu2015Multi}. Each layer takes all preceding feature maps as input. The transition down unit has a batch normalization layer, a 3$\times$3$\times$3 convolution layer with the ReLU activation function and a 2$\times$2$\times$2 max-pooling layer with the stride of 2. The number of filters in 3D convolution layers is a fixed value in each PRN block. This value is initially 32 and increases from the top PRN block to the bottom one with a step of 16.
The decoder has an aggressive deconvolutional path, which is similar to the encoder path except that (1) all blocks are in an inverse sequence, and (2) the last 3$\times$3$\times$3 convolution layer has a sigmoid activation function, instead of ReLU. The feature maps of each input path are also concatenated and copied to the corresponding location on the decoder path. The output of the decoder was binarized with a threshold of 0.8, and the connected volumes inside the output are detected as nodule candidates.

\subsection{HSN-Based Differentiation of Nodules from Non-Nodules}
\begin{figure}
\setlength{\intextsep}{0pt plus 0pt minus 1pt}
\setlength{\belowdisplayskip}{-1pt}
\setlength{\abovedisplayskip}{-1pt}
\setlength{\abovecaptionskip}{-1pt}
\setlength{\belowcaptionskip}{-1pt}
\centering
\includegraphics[width=0.90\textwidth]{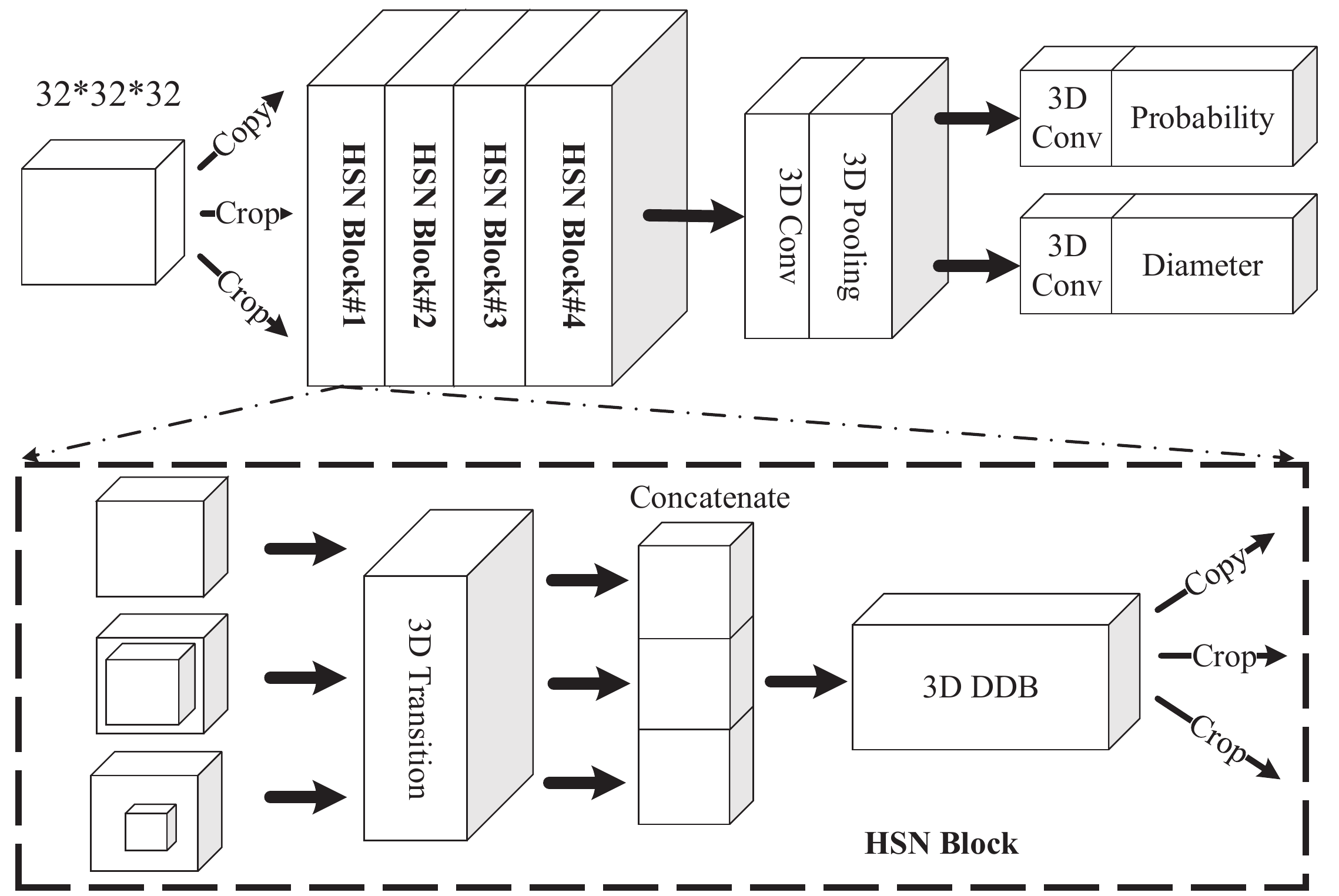}
\caption{Structure of the proposed HSN model}
\label{fig3}
\end{figure}
The proposed HSN model (see Fig. \ref{fig3}) consists of four subsequent HSN blocks, a 3$\times$3$\times$3 convolutional layer, a 2$\times$2$\times$2 max-pooling layer and two 1$\times$1$\times$1 convolutional layers. Each HSN block takes three volumetric patches as input, which are cropped on the chest CT scan according to the center of the detected nodule candidate. These patches, whose original sizes are 32$\times$32$\times$32, 24$\times$24$\times$24 and 16$\times$16$\times$16, respectively, are then resized to 32$\times$32$\times$32 by padding 0 and fed to the 3D transition layer. Three outputs of the transition layer are concatenated as the input of a 3D DDB for further processing. The regression layer uses the mean absolute error as the loss function and replaces the sigmoid function with a linear function. The structures of 3D DDB and 3D transition layers are the same to those in PRN. Similarly, the number of filters is initialized to 32 and increases with a step of 32. To improve its regularization ability, dropout with a probability of 0.5 is implemented on the output of max-pooling layers. The last two 1$\times$1$\times$1 convolutional layers are parallel output layers, which classify the candidate nodule and estimate the diameter of nodules, respectively.

\subsection{Training and testing}

To alleviate the overfitting of our models, we performed data augmentation to increase the number of positive samples in training and testing. Specifically, we generated five augmented copies for each nodule via rotating the nodule along the Z axis with 0°, 90°, 180° and 270° and left-right flipping it.
At the training stage, we cropped a 32$\times$32$\times$32 patch around each nodule as a positive sample and randomly cropped negative samples, including pleura regions, trachea tissues and vessels, to ensure that the ratio of positive samples, together five folds of augmented copies, and negative samples is 1:10. We initialized the biases in each model as 0 and initialized other parameters by sampling from a standard normal distribution. We chose the mini-batch stochastic gradient descent method with a batch size of 16 and a momentum of 0.9 as the optimizer. We set the learning rate to 0.01 and decreased it to 0.001 after 20 epochs. We randomly selected 20\% of the training data to form a validation set and stopped the training when the accuracy on the validation set had not improved for more than 20 epochs.
At the testing stage, we slid a 32$\times$32$\times$32 window over the lung and fed each volumetric patch inside the window to the trained PRN to detect nodule candidates. Then we cropped three concentric patches according to the center of the detected nodule candidate and fed them, together five folds of augmented copies, to the trained HSN for true nodule identification and diameter estimation. The genuineness of the nodule candidate is determined by majority voting and its diameter is determined by averaging the estimated values.
\section{Experiments and Results}

\begin{figure}
  \centering
  \includegraphics[width=3in]{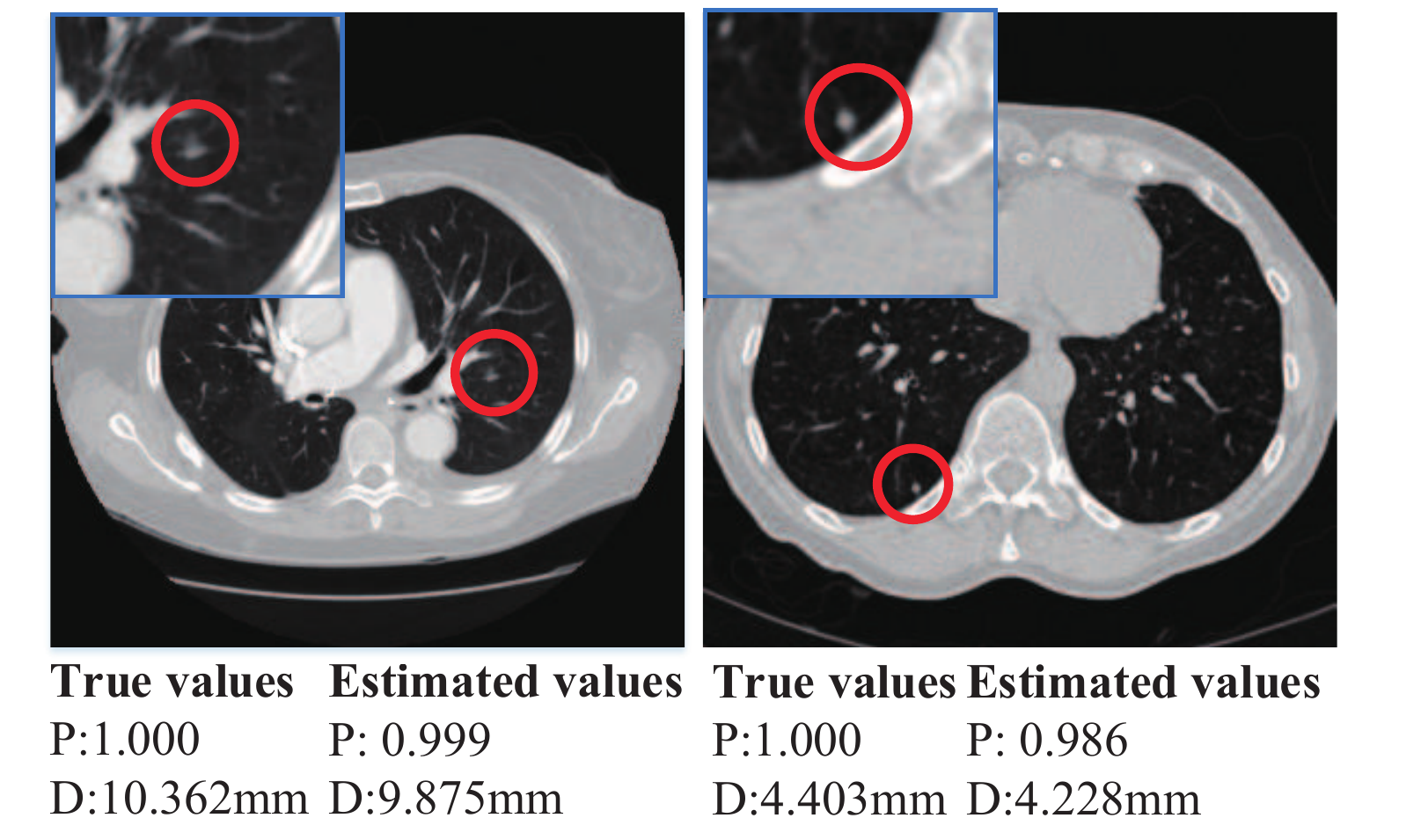}
  \caption{\label{fig:4} Sample images of estimated results}
\end{figure}

We evaluated the proposed pulmonary nodule detection algorithm on the LUNA16 dataset using 10-fold cross-validation. Fig. \ref{fig:4} shows two example cases, where “P” represents the probability of being a genuine nodule, and “D” is the diameter of the nodule. It reveals that our method is able to detect small nodules, whose diameter is less than 5mm, and can estimate nodule diameter accurately. Moreover, the average difference between the detected diameter and true diameter of detection nodules is 1.34 mm.
\subsubsection{Comparison to challenge records.}Next, we compared the performance of our method to the top five records on the LUNA16 Challenge leaderboard in Table \ref{table1}, where the score is calculated as the average sensitivity at seven operating points of the FROC curve: 1/8, 1/4, 1/2, 1, 2, 4, and 8 FPs/scan, and the 95\% confidence interval is computed using bootstrapping with 1,000 bootstraps. It shows that our method achieved a score of 0.958, which is higher than the current best challenge record and, to our knowledge, is so far the best performance ever achieved on this dataset.

\begin{table}[]
\centering
\caption{PARTIAL RESULTS OF LUNA16 CHALLENGE LEADERBOARD}
\label{table1}
\begin{tabular}{clc}
\hline
Rank & \multicolumn{1}{c}{Team}   & Score \\ \hline
1    & Our method                 & 0.958 \\
2    & PAtech (PA\_tech)          & 0.951 \\
3    & JianPeiCAD (weiyixie)      & 0.950 \\
4    & LUNA16FONOVACAD(zxp774747) & 0.947 \\
5    & iFLYTEK-MIG (yinbaocai)    & 0.941 \\
6    & iDST-VC (chenjx1005)       & 0.922 \\ \hline
\end{tabular}
\end{table}
\subsubsection{Computational Complexity }
It took about 18 and 23 hours to train the proposed PRN and HSN models, respectively, and took about three minutes to detect pulmonary nodules on a chest CT scan (Intel Xeon E5-2640 V4 CPU, NVIDIA Titan X GPU, 512 GB RAM, Ubuntu 16.04, Python 3.5.2, Tensorflow 1.5.0 and Keras 2.1.3). Since the training process that can be performed offline, our pulmonary nodule detection method can analysis one scan in three minutes and could be used in a routine clinical workflow.
\section{Conclusions }
This paper proposes the PRN model to detect pulmonary nodule candidates on chest CT scans and the HSN model to identify genuine nodules form those candidates. Our results show that our pulmonary nodule detection method achieved a score of 0.958 on the LUNA16 dataset, which is higher than the current best challenge record and, to our knowledge, is so far the best performance ever achieved on this dataset.
\bibliographystyle{splncs03}
\bibliography{References}

\begin{thebibliography}{10}
\providecommand{\url}[1]{\texttt{#1}}
\providecommand{\urlprefix}{URL }

\bibitem{Aaa2016Validation}
Aaa, S., Traverso, A., De, B.T., Msn, B., Cvd, B., Cerello, P., Chen, H., Dou,
  Q., Fantacci, M.E., Geurts, B.: Validation, comparison, and combination of
  algorithms for automatic detection of pulmonary nodules in computed
  tomography images: The luna16 challenge. Med. Image Anal.  42, ~1 (2016)

\bibitem{Baldwin2015Prediction}
Baldwin, D.R.: Prediction of risk of lung cancer in populations and in
  pulmonary nodules: Significant progress to drive changes in paradigms. Lung
  Cancer  89(1),  1--3 (2015)

\bibitem{Callister2016How}
Callister, M.E., Baldwin, D.R.: How should pulmonary nodules be optimally
  investigated and managed? Lung Cancer  91, ~48 (2016)

\bibitem{Dou2017Automated}
Dou, Q., Chen, H., Jin, Y., Lin, H., Qin, J., Heng, P.A.: Automated pulmonary
  nodule detection via 3d convnets with online sample filtering and hybrid-loss
  residual learning. In: MICCAI. pp. 630--638 (2017)

\bibitem{Ferlay2015Cancer}
Ferlay, J., Soerjomataram, I., Dikshit, R., Eser, S., Mathers, C., Rebelo, M.,
  Parkin, D.M., Forman, D., Bray, F.: Cancer incidence and mortality worldwide:
  sources, methods and major patterns in globocan 2012. Int. J. Cancer  136(5),
   E359 (2015)

\bibitem{Hamidian20173D}
Hamidian, S., Sahiner, B., Petrick, N., Pezeshk, A.: 3d convolutional neural
  network for automatic detection of lung nodules in chest ct. Proc. SPIE
  10134,  1013409 (2017)

\bibitem{Huang2017Densely}
Huang, G., Liu, Z., Maaten, L.V.D., Weinberger, K.Q.: Densely connected
  convolutional networks. In: CVPR (2017)

\bibitem{Kohan2013N}
Kohan, A.A., Kolthammer, J.A., Vercher-Conejero, J.L., Rubbert, C., Partovi,
  S., Jones, R., Herrmann, K.A., Faulhaber, P.: N staging of lung cancer
  patients with pet/mri using a three-segment model attenuation correction
  algorithm: Initial experience. Eur. Radiol.  23(11),  3161 (2013)

\bibitem{Nithila2016Automatic}
Nithila, E.E., Kumar, S.S.: Automatic detection of solitary pulmonary nodules
  using swarm intelligence optimized neural networks on ct images. J. Eng. Sci.
  Tech.  20(3) (2016)

\bibitem{Ohtsu1979A}
Ohtsu, N.: A threshold selection method from gray-level histograms. IEEE Trans.
  Syst. Man Cybern.  9(1),  62--66 (1979)

\bibitem{Valente2016Automatic}
Valente, I.R., Cortez, P.C., Neto, E.C., Soares, J.M., de~Albuquerque, V.H.,
  Tavares, J.M.: Automatic 3d pulmonary nodule detection in ct images: A
  survey. Comput. Methods Prog. Biomed.  124(C),  91--107 (2016)

\bibitem{Wu2016Correlation}
Wu, P., Xia, K., Yu, H.: Correlation coefficient based supervised locally
  linear embedding for pulmonary nodule recognition. Comput. Methods Prog.
  Biomed.  136, ~97 (2016)

\bibitem{Yu2015Multi}
Yu, F., Koltun, V.: Multi-scale context aggregation by dilated convolutions.
  arXiv:1511.07122  (2016)

\end{thebibliography}

\end{document}